\documentclass[times,twocolumn,final]{elsarticle}

\usepackage{cag}
\usepackage{framed,multirow}
\usepackage{textcomp}
\usepackage{amssymb}
\usepackage{latexsym}
\usepackage{multirow,gensymb,float}
\usepackage{amsmath}

\usepackage{url}
\usepackage{xcolor}
\definecolor{newcolor}{rgb}{.8,.349,.1}

\usepackage{hyperref}

\journal{ARXIV}

\begin{document}

\verso{Preprint Submitted for review}

\begin{frontmatter}

\title{Benchmarks for Industrial Inspection Based on Structured Light}%

\author[1,3]{Yuping  \snm{Ye}}
\cortext[cor1]{Corresponding author: Zhan Song is with Shenzhen Institute of Advanced Technology, Chinese Academy of Sciences, Shenzhen, China}
\emailauthor{yp.ye@siat.ac.cn}{Yuping Ye}
\ead{example@email.com}
    \author[2]{Siyuan \snm{Chen}}
 \emailauthor{sychen@gci3d.com}{Siyuan Chen}
    
\author[1,4]{Zhan \snm{Song}\corref{cor1}}
\emailauthor{zhan.song@siat.ac.cn}{Zhan Song}

\address[1]{Shenzhen Institute of Advanced Technology, Chinese Academy of Sciences, Shenzhen, China}
\address[2]{ GC Inovation Co. Ltd, Shenzhen, China}
\address[3]{University of Chinese Academy of Sciences, Beijing, China}
\address[4]{The Chinese University of Hong Kong, Hong Kong}

\received{\today}

\begin{abstract}
Robustness and accuracy are two critical metrics for industrial inspection. 
In this paper, we propose benchmarks which can evaluate the structured light method's performance .
Our evaluation metric was learning from a lot of inspection tasks from the factories.
The metric we proposed consists of four detailed criteria such as flatness, length, height and sphericity.
Then we can judge whether the structured light method/device can be applied to specified inspection task by our evaluation metric quickly.
A structured light device built for TypeC pin needles inspection performance is evaluated via our metrics in the final experimental section.

\end{abstract}

\begin{keyword}
\KWD Strucutre Light\sep Industrial Inspection\sep 3D Vision\sep 3D Reconstruction

\end{keyword}

\end{frontmatter}

\section{Introduction}
With the fast development of manufacturing, a lot of heavy and tedious work can be done by machines automatically, which improves the efficiency significantly.
By comparison, the inspection as the last and crucial step in making products, especially for 3C products (Computer, Communication, Consumer Electronic products)  still involved much labour.
With the development trend of device miniaturization, more and more inspection work cannot be completed by simple labour without a specialized and complicated tool.
Take the mobile phone as an instance; mobile phones' size decreases each year dramatically, making more stringent requirements for inspection tools.
Recently,  The famous contract manufacturer  Foxconn Electronics is trying to introduce structured light based inspection machine for fault detection.
As one of the most essential noncontact optical 3D measurement methods, structure light has been broadly used in a variety of applications such as virtual reality, animation, and industrial inspection.

  \begin{figure}[!t]
	\centering
	\includegraphics[width=\columnwidth]{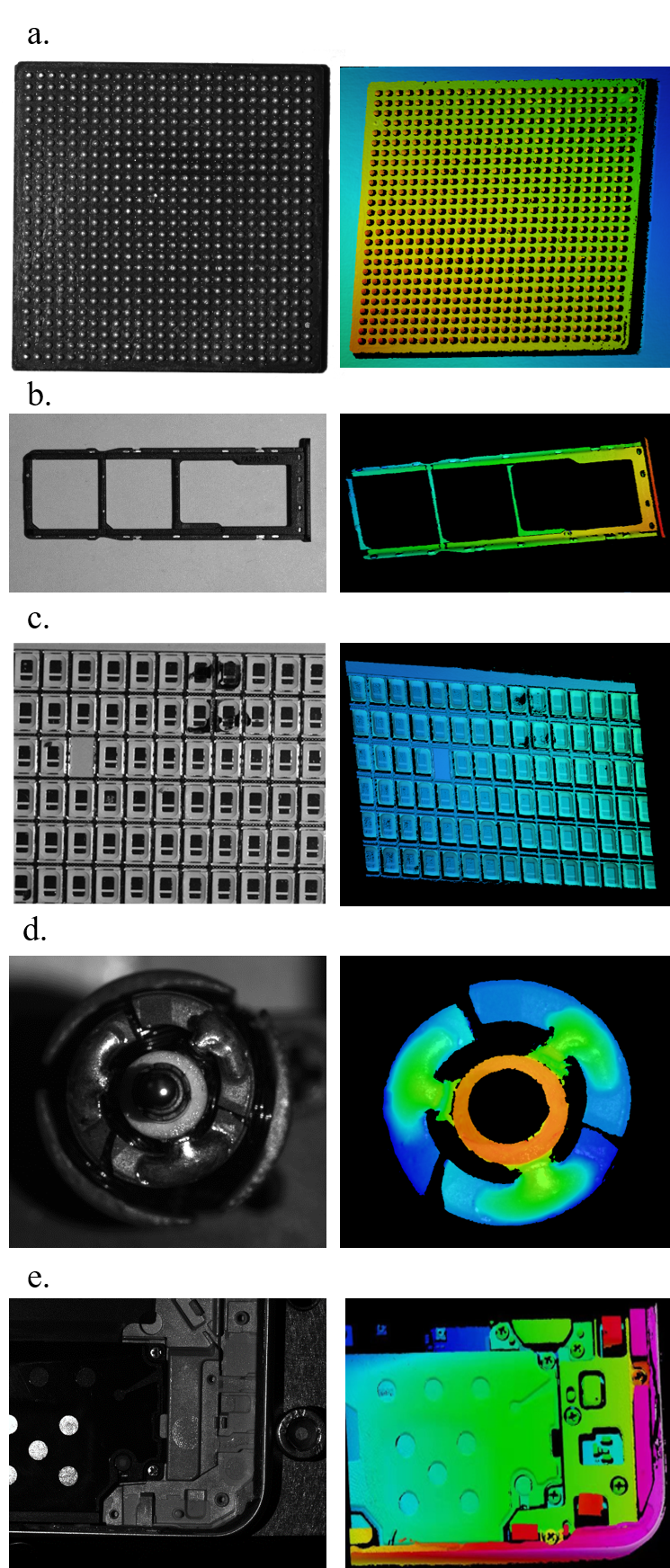}
	\caption{Several inspection cases from assemble line.
		 a). shows coplanarity inspection of BGA solder balls,   b). shows SIM card tray's edge height measurement, c) shows LED's defect detection
		d) shows solder paste area measurement on an small motor , e) shows the screw holes positioning and edge coplanarity checking etc on mobile phone.}
	\label{sampleuse}
\end{figure}

 A typical structured light system consists of one camera and one light emitter(projector/laser emitter).  
The light emitter project the coded pattern onto the target object. The camera is then adopted to capture the illuminated target object; finally, the decoding algorithm is utilized for these images to obtain geometrical information of the target object.
Although structured light technology has a research history of more than forty years, it is only in recent years that structured light technology being used to inspect small and microscale devices.
This is mainly caused by high-resolution cameras' popularity in recent years and the miniaturization trend of devices.
Due to the characteristic that structured light easy to build and integrate into the detection equipment, it has gained more and more attention from the industry.
Several inspection cases from production  line based on structure light technology are presented in  Fig. \ref{sampleuse}: Left column shows the texture image of the workpieces need to inspect, the right column shows the 3D model obtain by structured light technology.

The inspection method/device which deploys on an assembly line must be very robust and accurate.
In scientific research, researchers always pay more attention to accuracy than robustness and use different metrics to evaluate their proposed reconstruction methods.
These situations make it difficult for engineers to select a suitable structured light method/device for inspection tasks.
To bridge the gap between industrial inspection and scientific research, we propose evaluation metrics for measuring the structured light method/device's performance.
Our evaluation metric was learning from a lot of inspection cases from the assembly line.
We can judge whether the structured light method/device can be applied to specified inspection task by our evaluation metric.

The organization of this paper is as follows. 
Previous work related to structured light and industrial inspection is briefly reviewed in Sec 2. 
Preliminary knowledge of structured light is presented in Sec 3.
A detailed introduction to our benchmarks for industrial inspection based on structured light is proposed in Sec 4. 
Experimental results are provided and discussed in Sec 5.

\section{Related Work}

According to the codeword of patterns adopted by structured light, we can categorize projector structure light into three classes: direct encoding, spatial-multiplexing, and time-multiplexing. Direct encoding structured light methods directly decode each pixel by its intensity/colour, so it needs few patterns. Spatial-multiplexing structured light methods only require one pattern image, and so the codeword of a specific pixel is decoded from several pixels.  Unlike spatial-multiplexing groups, time-multiplexing methods need to project several patterns sequentially. As a consequence, the codeword for a given pixel is formed by a sequence of patterns. The advantage of time-multiplexing methods is obvious; it can achieve dense and accurate results and is robust against the surface texture.
Caused the industrial inspection's high requirement on accuracy and robustness, most of the inspection methods are based on the time-multiplexing.
So we will focus on surveying the time-multiplexing projector structured light in this section.

Due to strutured light's  long  research history, there have been a lot of  researchers  \cite{xu2020line,RN47,RN49,RN50,RN52,RN51,angelsky2020structured} provided a review of advances in structured light scanning at different times. Most efforts on the structured light have been devoted to the following topics: laser center extracting\cite{sun2015robust,ALTEMEEMY2021106897}, accuracy \cite{RN53,RN54}, reflective object \cite{RN56,RN57}, calibration \cite{RN61,deetjen2018automated,klemm2017high}, and code strategy \cite{RN58,RN59,RN60} etc. 
For the sake of the laser structured light's simplicity, it has been well studied.
In \cite{ko2018development},  a colour 3D scanner based on the laser structured-light imaging method
that can simultaneously acquire 3D shape data, and colour of a target object is presented.
Sun \cite{sun2015robust} put forward a  closed-form solution for extracting the laser stripe centre.
In \cite{ackerman2008methods},  Methods and systems for laser-based real-time structured light depth extraction is desecribed.
Numerous work has contributed to the time-multiplexing structured light. 
One of the most widely used time-multiplexing coding strategies is the Gray code method. In Gray code methods, each codeword only has one different bit different between random alternate code values, and all bits are with the same intensity, so it is robust and straightforward. Because the Gray code can remove periodic ambiguities whereas line or phase-shifting code achieves sub-pixel correspondence mapping, many authors have combined the Gray code with line-shifting \cite{ RN93, zheng2010structured, yang2008robust}, or phase-shifting \cite{wu2019high,sansoni1999three, yu2016unequal} stripes. In \cite{ RN93}, the author encoded with Gray code stripe and decoded with stripe edge acquired by sub-pixel technology instead of pixel centre, and also corrected the error caused by dividing projecting angle irregularly by even-width encoding stripe.  Wu et al. \cite{wu2019high} put forward a method which combines the Gray code with phase-shifting stripes to overcome the phase unwrapping errors in the high-speed measurement. 
A structure light method \cite{cai2020structured} without phase unwrapping is presented to adapt to the measurement environment changes.
Due to the structured light's simplicity and accurate, many off-the-shelf products \cite{li2020inspection,cofer2007method,chang2020three} based on structured light have been widely used.

Compared with structured light, the research of industrial inspection based on structured light is relatively rare.
In \cite{wei2011coplanarity}, the authors put forward an inspection method based on laser structured light for checking the coplanarity of the BGA chips. 
Recently, BGA chips' coplanarity checking is often based on projector structured light methods.
A laser structured light sensor \cite{han2020structured} is developed to achieve on-line weld bead measurement and weld quality inspection.
In \cite{drouin2020active}, the authors show serveral inpsection cases based on structured light.
A systematic algorithm \cite{LIU2008617} of generating optimal inspection planning based on CAD-directed model is proposed for laser-guided measuring robot (LGMR) to inspect free-form surfaces.
The authors \cite{HUANG20151} surveys various inspection systems based on 2D optical images/videos in the semiconductor industry.
With the fast development of deep learning technology in recent years, researchers also try to introduce deep learning for inspection.
A deep residual network With adaptively parametric rectified linear units is proposed by Zhao et al. \cite{zhao2020deep} for fault diagnosis.
Although methods based on deep learning may gain plausible results, its' general ability for varies inspection tasks is weak, and the training process is time-consuming.

The linkage between the state-of-art structured light methods and the inspection applications is weak.
So we proposed benchmarks for engineers to evaluate state-of-art structured light methods for industrial inspection applications.

\section{Structure Light}
As the above mentioned, the most commonly used  structured light in industrial inspection is laser and time-multiplexing projector structured light.
In this section, we will illustrate the mathmatics behind these methods.
\subsection{Camera model}

\begin{figure}[H]
	\centering
	\includegraphics[width=\columnwidth]{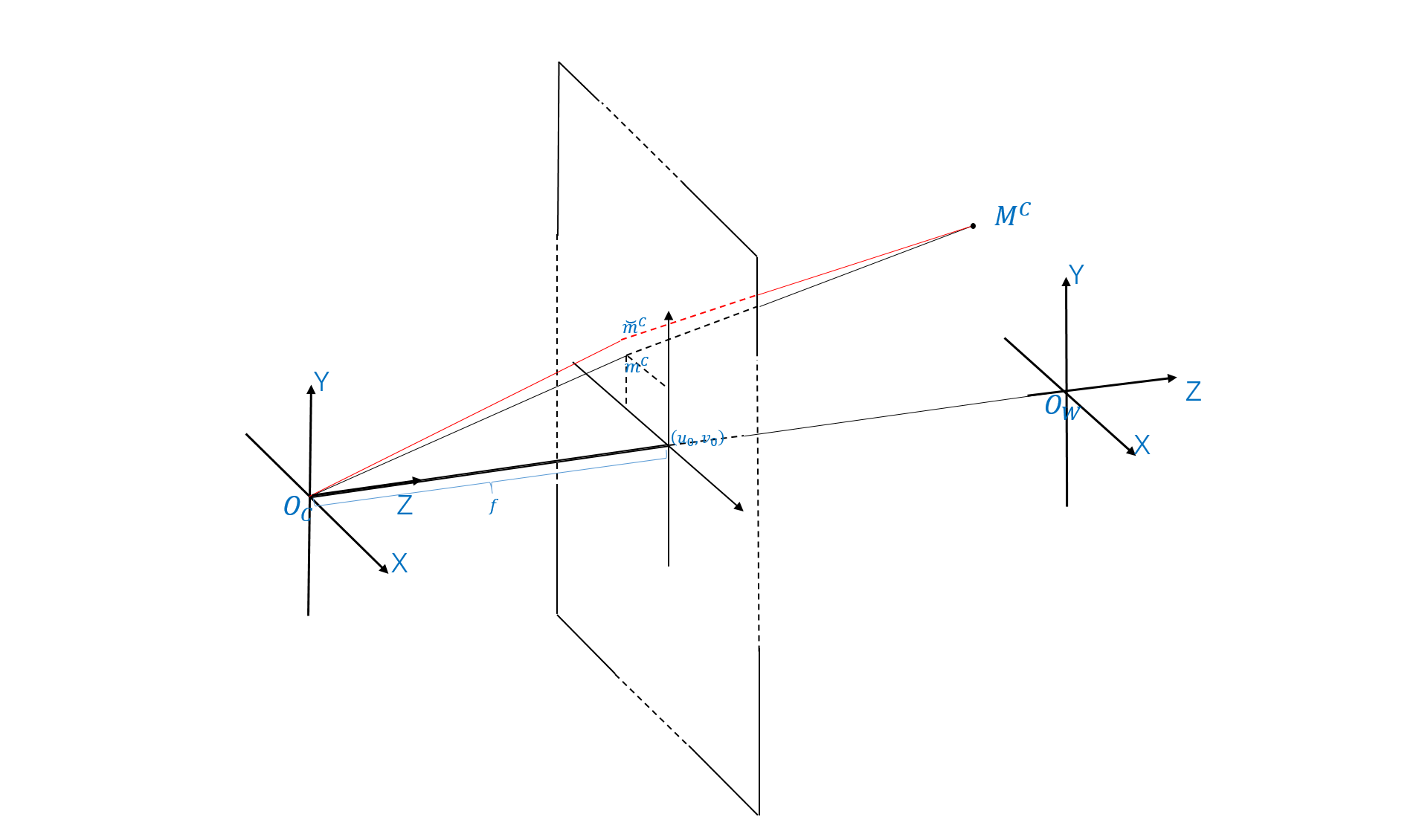}
	\caption{Camera's geometric model.}
	\label{cameramodel}
\end{figure}
Camera is an indispensable device in computer vision.
The geometric model of a typical camera is represented in Fig. \ref{cameramodel}. 
Based on the pinhole projection model, one 3D point with respect to the camera coordinate system $M^{C}={\begin{bmatrix}
		X^{C}&Y^{C}&Z^{C}
\end{bmatrix}}$ will be project onto the camera plane at  $m^{C}={\begin{bmatrix}
u^{C}&v^{C}
\end{bmatrix}}$.
And this can be  expressed by the following equation:
\begin{equation}
	s{\begin{bmatrix}			
			m^{C} \\ 1			
	\end{bmatrix}}=\underbrace{
		\begin{bmatrix}f_u^{C} &\gamma_u^{C} &u_0^{C}\\ 
			0          &f_v^{C}  &v_0^{C} \\
			0   &0 &1
		\end{bmatrix}
	}_{K^{C}}
	{
		\begin{bmatrix}
			I_3&\vec{0}
		\end{bmatrix}
	}
	{
		\begin{bmatrix}
			M^{C}\\
			1
		\end{bmatrix}
	}
	\label{eq:eq1}
\end{equation}
where the subscript $C$ denoete it's camera parameters, $K^{C}$ known as the intrinsic matrix is composed of focal length $f_u^{C}$  ,$f_v^{C}$  , principal point $u_0^{C}$ ,$v_0^{C}$ , and skew coefficient $\gamma_u^{C}$.
Let  $\bar{M}^{C}$ be the normalized image projection point:
\begin{equation}
	\bar{M}^{C}=\begin{bmatrix}
		X^{C}/Z^{C}\\
		Y^{C}/Z^{C}
	\end{bmatrix}=
	\begin{bmatrix}
		x^{C}\\
		y^{C}
	\end{bmatrix}
	\label{eq:eq2}
\end{equation}
So the Equation (\ref{eq:eq1}) can be rewritten as follow:
\begin{equation}
	{\begin{bmatrix}
			
			m^{C} \\ 1
			
	\end{bmatrix}}=
	K^{C}
	{\begin{bmatrix}
			
			\bar{M}^{C} \\ 1
			
	\end{bmatrix}}
	\label{eq:eq3}
\end{equation}
We can transform the point $M^{C}$ in the camera coordinate system into the world coordinate system  $M^W$ . So we can correlate   $M^C$ and $M^W$   as the following equation:
\begin{equation}
	{\begin{bmatrix}
			
			M^{C} \\ 1
			
	\end{bmatrix}}=
	{\begin{bmatrix}
			
			R^C &T^C\\
			0 &1
			
	\end{bmatrix}}
	{\begin{bmatrix}
			
			\bar{M}^{W} \\ 1
			
	\end{bmatrix}}
	\label{eq:eq4}
\end{equation}
where $R^C$  and $T^C$  indicate the rotation matrix and transition vector between   the camera and the world coordinate system.
Due to the     manufacturing and assembly's limitation,  the simple pinhole model failed in many cases.
So we have to take  the radial distortions $k_1^{C}$,$k_2^{C}$,$k_3^{C}$  and tangential distortions $p_1^{C}$ ,$p_2^{C}$  into consideration. Due to the existence of the distortions, the real normalized image projection point  $\hat{M}^{C}$ may have deviation with  :
\begin{equation}
	\hat{M}^{C}=\Upsilon(\bar{M}^{C})+\Delta(\bar{M}^{C})
	\label{eq:eq5}
\end{equation}
where $\Upsilon(\bar{M}^{C})$  and $\Delta(\bar{M}^{C})$  refers to radial and tangential distortion vector respectively. 
Let $(r^{C})^2=(x^{C})^2+(y^{C})^2$ , and the radial and tangential distortion vector can be expressed as:
\begin{equation}
	\begin{split}
		\Upsilon(\bar{M}^{C})=
		\bar{M}^{C}(1+k_1^{C}(r^{C})^2&+k_2^{C}(r^{C})^4\\
		&+k_3^{C}(r^{C})^6)
	\end{split}
	\label{eq:eq6}
\end{equation}

\begin{equation}
	\Delta(\bar{M}^{C})=
	\begin{bmatrix}
		
		2p_1^{C}x^{C}y^{C}+p2^{C}((r^{C})^2+2(x^{C})^2)\\ 
		2p_2^{C}x^{C}y^{C}+p1^{C}((r^{C})^2+2(y^{C})^2)
	\end{bmatrix}
	\label{eq:eq7}
\end{equation}

\begin{equation}
	{\begin{bmatrix}
			m^{C} \\ 1		
	\end{bmatrix}}=
	K^{C}
	{\begin{bmatrix}
			
			\hat{M}^{C} \\ 1
			
	\end{bmatrix}}
	\label{eq:eq8}
\end{equation}

And the most famous camera calibration method is proposed by Zhang \cite{zhang2000flexible}.

\subsection{Laser Structured Light}
Laser structured light as the most simple structured light which with high precision and easy to deploy.
Laser structured light is composed of a camera and a laser emitter. 
However, because the line laser can only obtain one point coordinate on the plane at a time, so in reality, the equipment using the line laser often needs to be matched with a displacement platform. This may reduce the line laser's accuracy and make it challenging to integrate into some detection equipment.
\begin{figure}[H]
	\centering
	\includegraphics[width=\columnwidth]{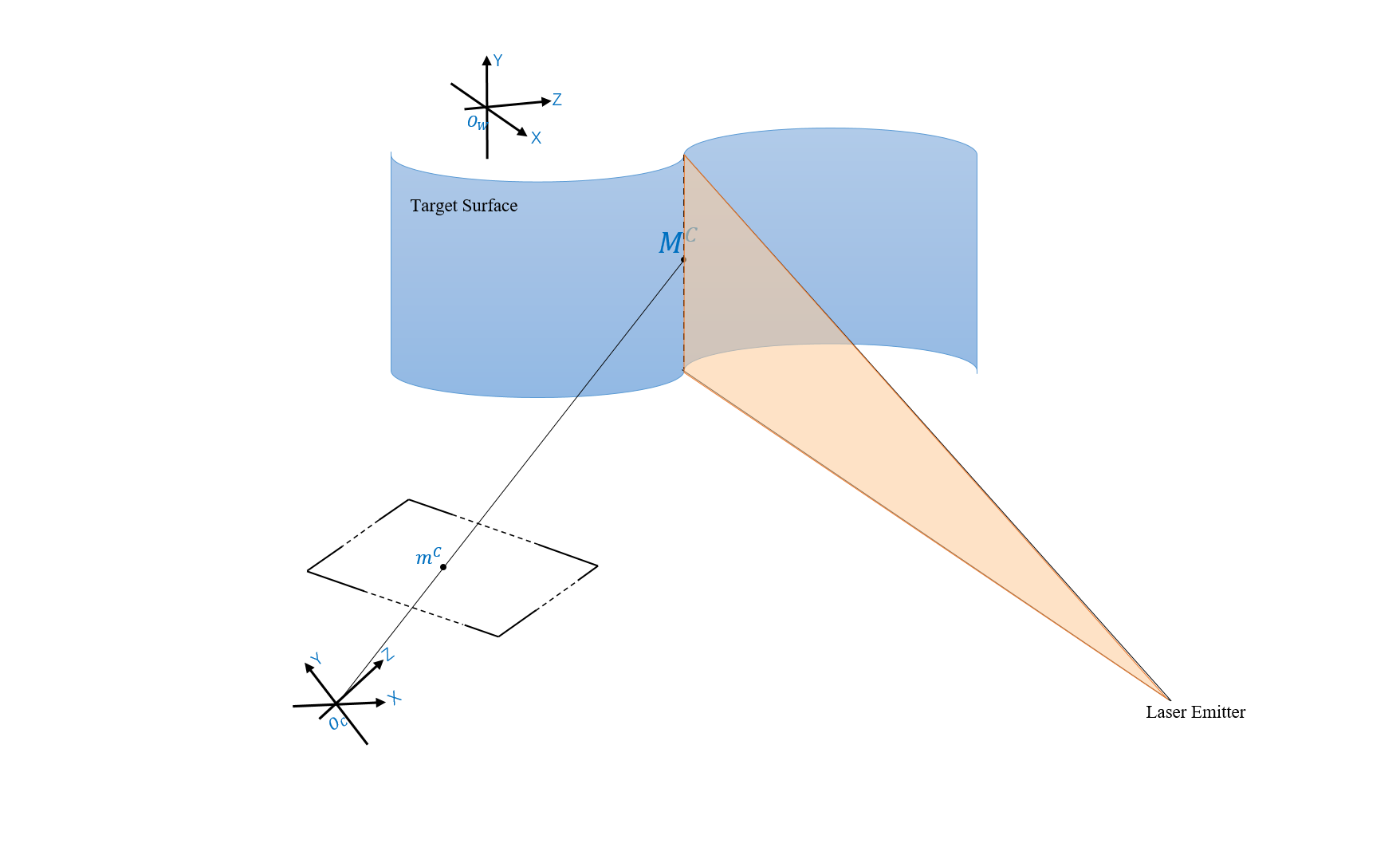}
	\caption{Geometric model of a typical laser structured light system.}
	\label{laserSL}
\end{figure}
The laser emitter emits a laser plane onto the target. Then,  the laser centre of the laser line can be extracted from the captured the illuminated image, final the point on the laser line can be calculated with simple triangulation method.
As the Fig. \ref{laserSL} shows,   a target surface 3D point  $\hat{M}^{C}$ on laser plane 
\begin{equation}
	LP:A'X+B'Y+C'Z+D'=0
	\label{eq:eq9}
\end{equation}
can be captured by camera .
The laser plane's cofficients $A',B',C',D'$ canbe calibrate according to Equation (\ref{eq:eq4}) .
Finally, we can get the Z coordinate as following:
\begin{equation}
Z=\frac{-D'}{A'(u-u_0^C)/f_u^C+B'(v-v_0^C)/f_v^C+C'}
	\label{eq:eq10}
\end{equation}

\subsection{Projector Structured Light}
Line lasers can only reconstruct one laser line and need the cooperation of the displacement platform.
Researchers have introduced the projector to replace the laser with a projector to rebuild the target surface at once.
\begin{figure}[H]
	\centering
	\includegraphics[width=\columnwidth]{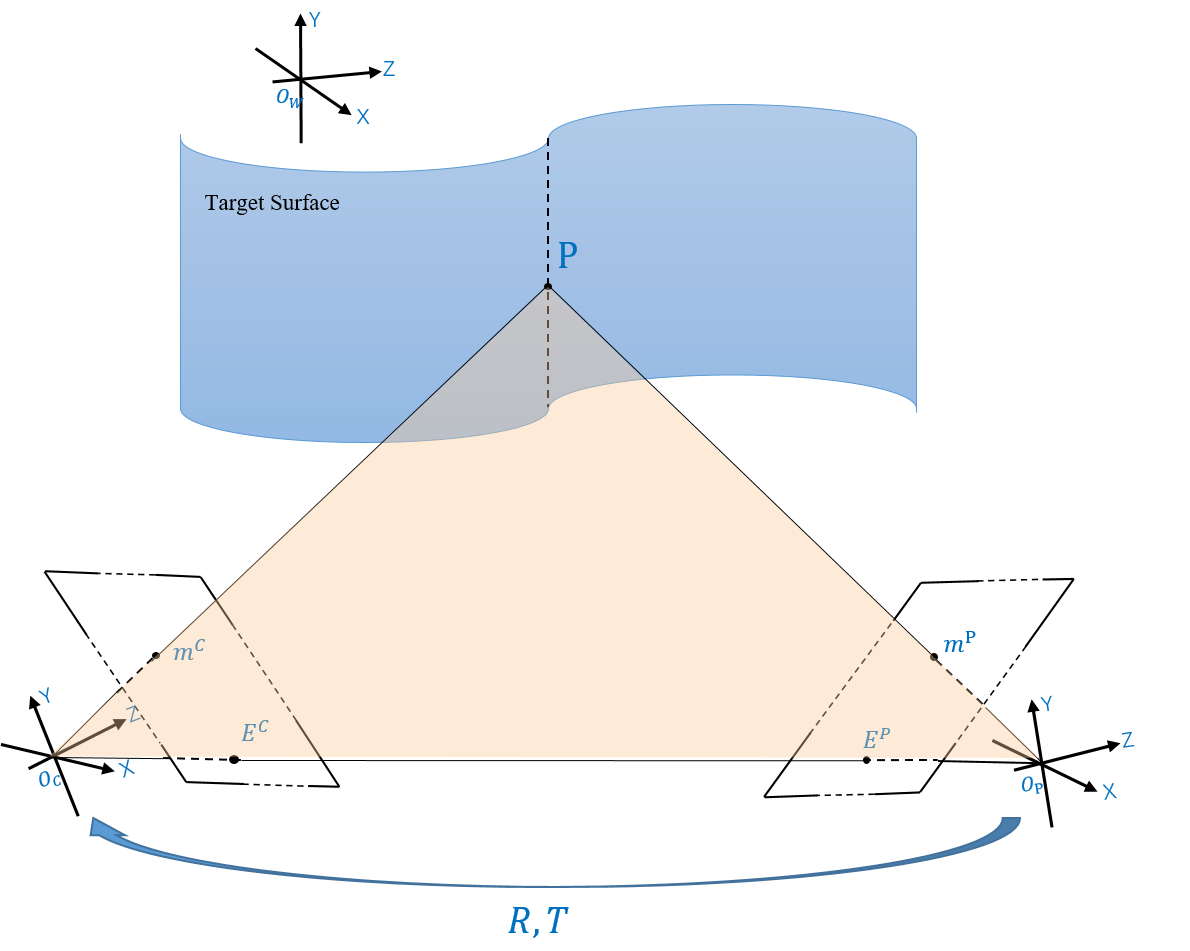}
	\caption{Geometric model of a typical projector structured light system}
	\label{projectorSL}
\end{figure}
As the Fig. \ref{projectorSL},  
The projector in the SLS system is often be treated as an inverse-light-path camera. 
So the projector also satisfied the Equation (1-8).
Unlike the binocular stereo vision, the SLS methods can easily find the correspondence by searching the features in the captured images. The workflow of the SLS system can be summarized into three steps. Firstly, patterns are projected onto the target and imaged by the camera synchronously. A decoding method is then adopted to extract the pattern feature from the captured images and establish the correspondences between the camera and the projector. Finally, the target's coordinates can be calculated via the triangulation formulation with the SLS system's calibration parameter.

We can transform the point $M^{C/P}$ in the camera or projector coordinate system into the world coordinate system. So we can correlate   $M^C$ and $M^P$   as the following Equation:
\begin{equation}
	{\begin{bmatrix}
			
			M^{P} \\ 1
			
	\end{bmatrix}}=
	{\begin{bmatrix}
			
			R &T\\
			0 &1
			
	\end{bmatrix}}
	{\begin{bmatrix}
			
			\bar{M}^{C} \\ 1
			
	\end{bmatrix}}
	\label{eq:eq11}
\end{equation}
where $R$  and $T$  indicate the rotation matrix and transition vector between the projector and the camera coordinate system.

Suppose that we have the extracted point $m^C$  and $m^P$  associated with the same spatial 3D point. 
According to Equation (\ref{eq:eq8}) ,  we can apply the intrinsic parameters and distortion parameters to normalize the $m^{C/P}$  , and get   $\hat{m}^{C/P}$ .As the Fig. \ref{projectorSL} shows, $O_CO_P$  is epipolar line, $E_1$  and $E_2$  are epipolar points, using the epipolar constraint:
\begin{equation}
	(\hat{m}^C)^T((K^P)^{-T}T(K^C)^{-1})\hat{m}^P=0
	\label{eq:eq12}
\end{equation}
Finally, we can obtain the depth value $Z^C$  by the triangulation algorithm as:
\begin{equation}
	Z^C=\frac
	{(R\hat{m}^C\hat{m}^P)(\hat{m}^CT)-\|\hat{m}^C\|^2(R\hat{m}^CT)}
	{\|R\hat{m}^C\|^2\|\hat{m}^P\|^2-(R\hat{m}^C\hat{m}^P)^2}
	\label{eq:eq13}
\end{equation}

\section{Evaluation Metric}

As the above mentioned, many 3D reconstruction methods based on the structured light is proposed, and the results presented by the authors seems plausible.
When it comes to the real situation that engineers try to employ these methods for industrial inspection, it often cost engineers a lot of time to measure whether the structured light approach meets the inspection task's standards.
In this section, we proposed an evaluation metric which can evaluate the structured light methods' performance on inspection task accurately.
The well-designed metric can significantly shorten engineers' time to choose a suitable structured light method for the specified inspection task.
Evaluation metrics we proposed can be an energy function for calibration or reconstruction like \cite{han2019accurate}.
Also, we hope this metric can help researchers evaluating their methods for industrial inspection in future.

The evaluation metric we proposed is tried out from numerous inspection tasks.
Inspection method based on structured light that passes our metric can be stable and effective in the assembly line.
The metric we proposed consists of four detailed criteria: flatness, length, height and sphericity.

In statistics, range, mean and standard variance are generally used to describe data distribution.
Range is equal to the maximum  error minus the minimum  error,
standard variance measures how distant or spread the numbers in a dataset are from the mean.
\begin{equation}
	\begin{aligned}
	R(\{X_i\})&=X_{max}-X_{min} \\
	\mu(\{X_i\})&=\frac{1}{N}\sum_{i=1}^{N}{X_i} \\
	\sigma(\{X_i\})&=\left(\frac{1}{N}\sum_{i=1}^{N}{(X_i-\mu(X_i))^2}\right)^{\frac{1}{2}} \\
     \end{aligned}
\end{equation}
In this paper, we also use the mean and standard variance to describe data distribution.
Therefore, unlike most papers which evaluate the error once, we need to conduct several tests on different height in the structured light device's depth of field.
The repeated test number we suggested is 50 or more.

\begin{figure}[H]
	\centering
	\includegraphics[width=\columnwidth]{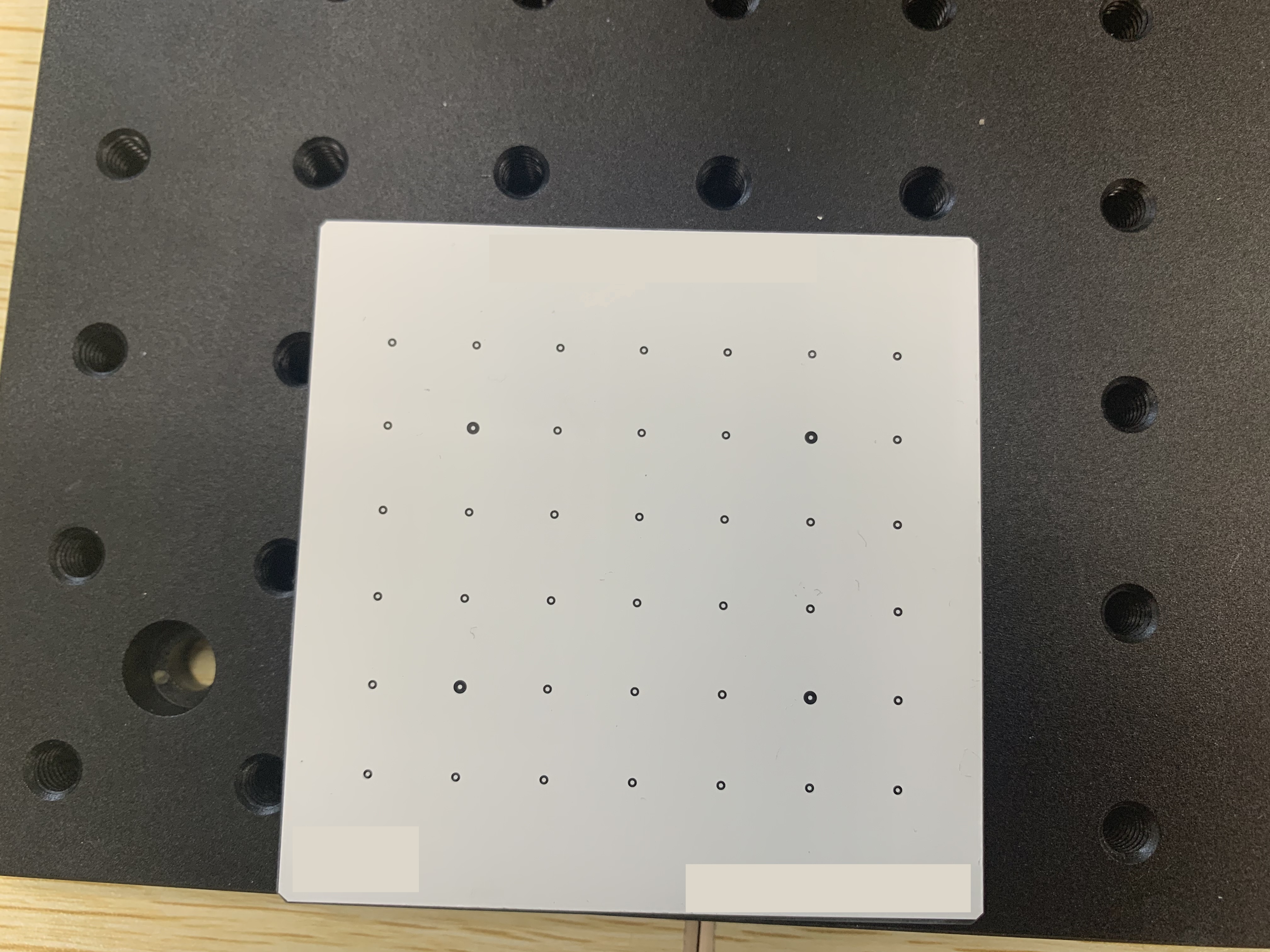}
	\caption{A precise flat with serveral hollow cricle markers is adopted for evluating the flatness and length.}
	\label{c1}
\end{figure}

Measuring the workpieces' flatness and length is the most common industrial inspection task.
To evaluate the structured light method's flatness, length, a precise flat is adopted, as shown in Fig. \ref{c1}. 
To eliminate the error caused by the thickness of  printing ink, several hollow circle markers are
uniformly printed on the flat with precise distance $l_c$ and angularity as shown in Fig. \ref{c1}.
For each hollow circle marker, we utilize the Hough circle detection algorithm \cite{kerbyson1995circle} to detect the marker's centre $P_i$.
For each of the two adjacent points $P_i,P_j$ in the horizontal or vertical direction, the length can be calculated and
denoted as $d_j$.
So the length error can caclulate as 
\begin{equation}
	E_d=d_j-l_c
\end{equation}
and the length metric can be represente as :
\begin{equation}
	\begin{aligned}
			&\mu(R(\{E_d\})),\,\,\sigma(R(\{E_d\})),\,\,\mu(\mu(\{E_d\})),\,\,\sigma(\mu(\{E_d\})),\\
		&\mu(\sigma(\{E_d\}))(Optional),\,\,\sigma(\sigma(\{E_d\}))(Optional)
	\end{aligned}
\label{eq:eq16}
\end{equation}

where $\mu(R(\{E_d\})),\,\,\sigma(R(\{E_d\}))$ denote the length error's range distribution,  $\mu(\mu(\{E_d\})),\,\,\sigma(\mu(\{E_d\}))$ for mean and $\mu(\sigma(\{E_d\})),\,\,\sigma(\sigma(\{E_d\}))$ for standard variance.
In a real inspection task, we often have little concern for the error's standard variance distribution.

To evaluate the flatness, we first use a  circle mask with a slightly larger radius than the actual radius for each hollow marker and remove these points according to Equation \ref{eq:eq1}.
The reference plane used for reconstruction can be viewed as a perfect plane.
Without considering the calibration errors and reconstruction errors, the planarity should be zero.
The least-square fitting approach is applied to fitting these points, and the plane function can be calculated as $AX+By+CZ+D=0$.
For each point's $P_i=(x_i,y_i,z_i)$ distance to the ideal plane canbe represented as:
\begin{equation}
	E_p=\frac{\|ax_i+By_i+Cz_i+D\|}{\left(A^2+B^2+c^2\right)^{\frac{1}{2}}}
\end{equation}

and the length metric can be represented as :
\begin{equation}
	\begin{aligned}
		&\mu(R(\{E_p\})),\,\,\sigma(R(\{E_p\})),\,\,\mu(\mu(\{E_p\})),\,\,\sigma(\mu(\{E_p\}))
	\end{aligned}
\label{eq:eq18}
\end{equation}

\begin{figure}[H]
	\centering
	\includegraphics[width=\columnwidth]{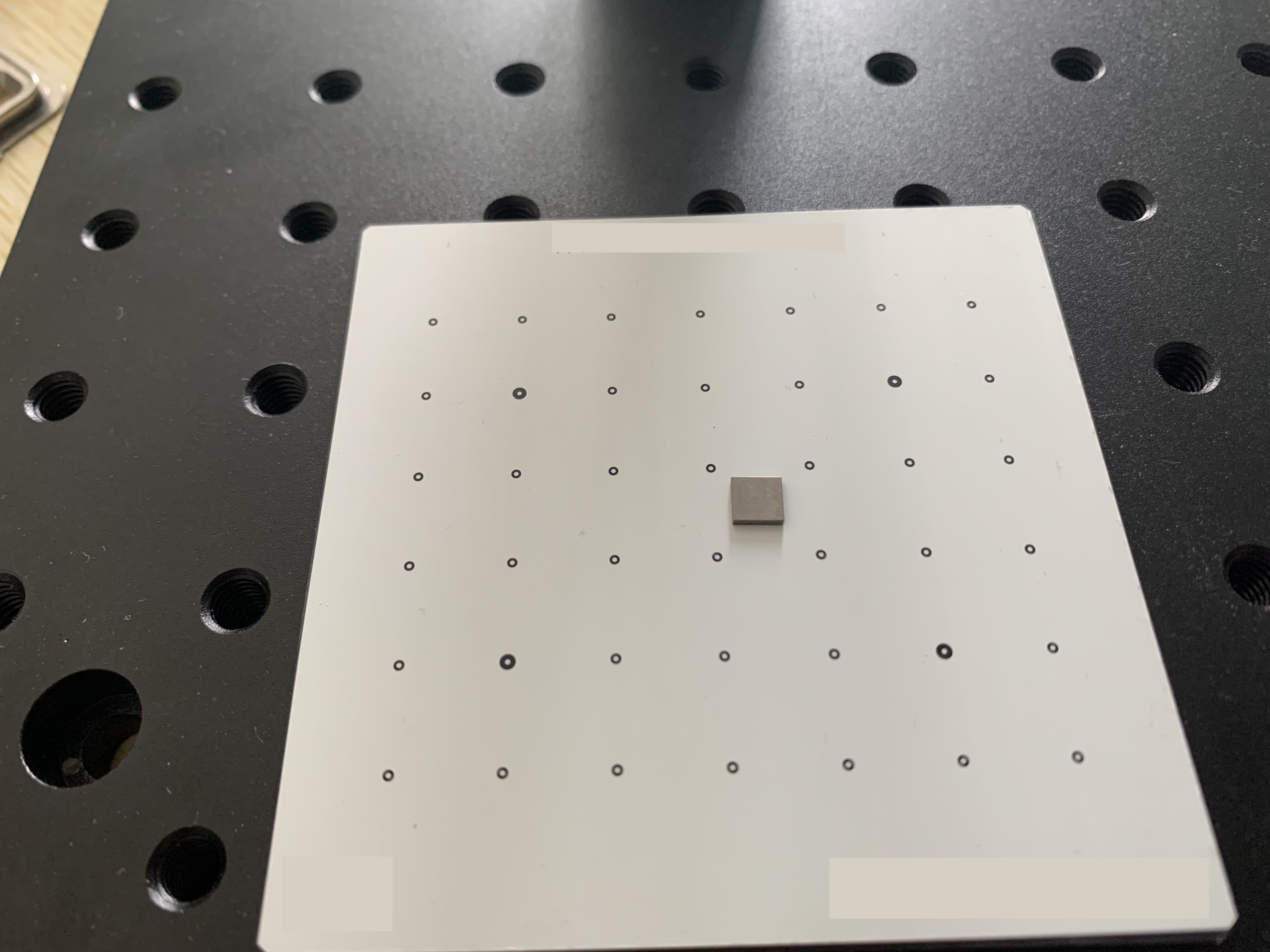}
	\caption{A gage block place on the flat to measure the height error.}
	\label{c2}
\end{figure}

Different from length error $E_d$, which always lager than the sensors' actual size of a pixel, the Z value of the structured light can have higher accuracy.
The most common inspection task is to calculate the height among workpiece's different parts.
As Fig.\ref{c2} show, we place a gauge  block with height $h_c$ onto the plane.
Like measuring the planarity did, we first use the flat's points without the gage block and hollow circles to get the plane function $AX+BY+CZ+D=0$; and then caculate each point $P_j=(x_j,y_j,z_i)$ distance to the fitting plane;
final the height error can be caculated as:
\begin{equation}
	E_h=\frac{\|ax_j+By_j+Cz_j+D\|}{\left(A^2+B^2+c^2\right)^{\frac{1}{2}}}-h_c
\end{equation}

So the height metric can be represented as :
\begin{equation}
	\begin{aligned}
		&\mu(R(\{E_h\})),\,\,\sigma(R(\{E_h\})),\,\,\mu(\mu(\{E_h\})),\,\,\sigma(\mu(\{E_h\}))
	\end{aligned}
\label{eq:eq20}
\end{equation}

\begin{figure}[H]
	\centering
	\includegraphics[width=\columnwidth]{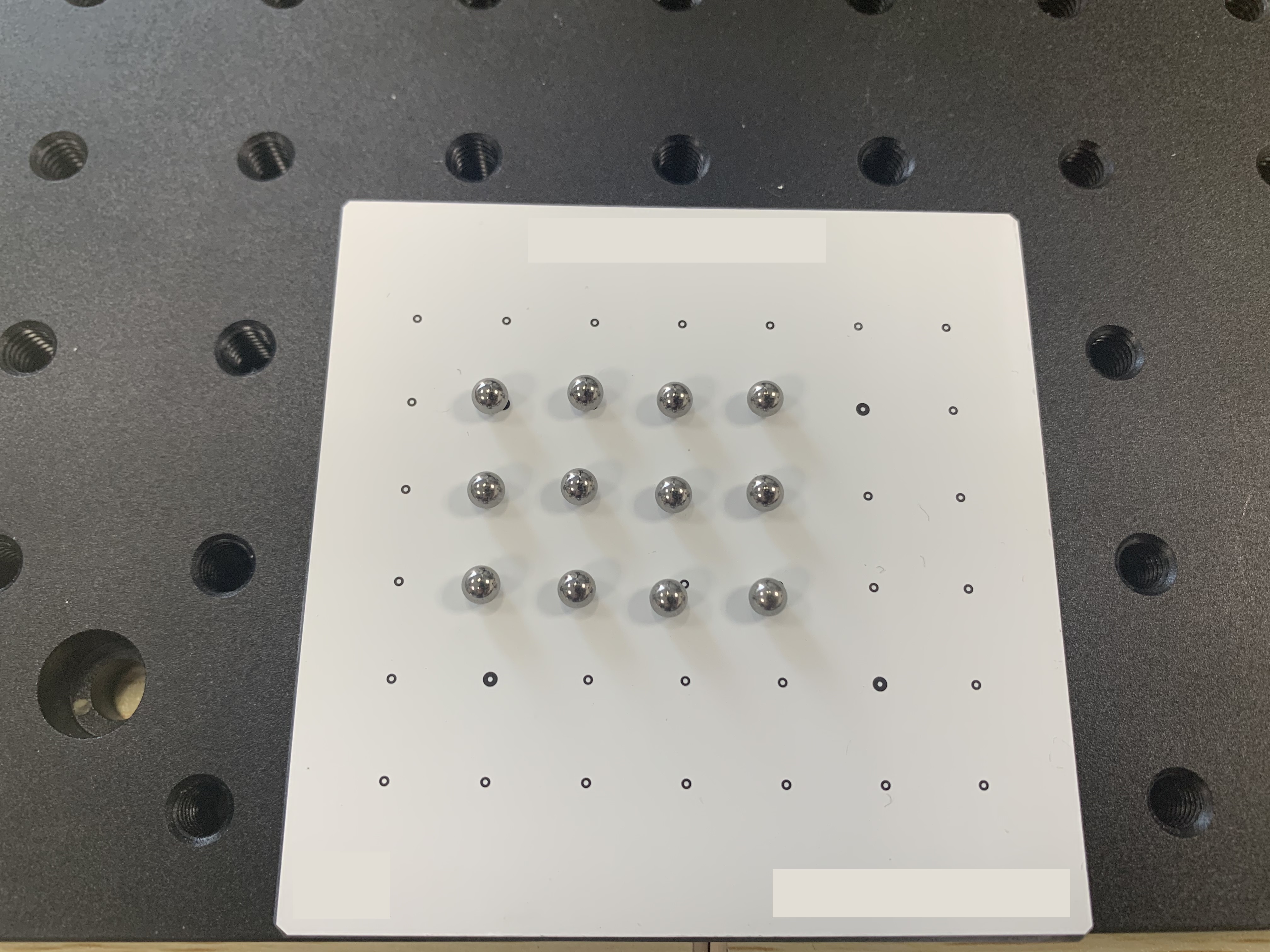}
	\caption{ Twelve tungsten steel balls are placed onto the flat to measure the sphericity.}
	\label{c3}
\end{figure}

Many algorithms, especially the off-the-shelf products, adopted many filter method to achieve better results, which can significantly promote the criteria such as planarity.
But overuse of the filter algorithms may lead to failure to inspect a workpiece with significant curvature.
To address this issue, we introduce the sphericity error.
As we know, the sphere is very sensitive to smooth filter methods.
As Fig. \ref{c3}, we place twelve tungsten steel balls with radius $r_c$ onto the plane, then capture its geometric information.
We can easily separate the balls' points from the plane via the height different, and apply the sphere fitting algorithm to get the ball centre and the radius $r_i$ of each ball.
The sphericity error can be caculated as:
\begin{equation}
	E_s=r_i-r_c
\end{equation}

and the sphericity metric can be represente as :
\begin{equation}
	\begin{aligned}
		&\mu(R(\{E_s\})),\,\,\sigma(R(\{E_s\})),\,\,\mu(\mu(\{E_s\})),\,\,\sigma(\mu(\{E_s\}))
	\end{aligned}
	\label{eq:eq22}
\end{equation}

As the above stated, our metric for inspection is composed with Equation \ref{eq:eq16},  \ref{eq:eq18}, \ref{eq:eq20} and  \ref{eq:eq22}.

\section{Experimental Results}

Traditionally,  inspection tasks are always raised firstly; then engineers seek to find a suitable solution according to the task's characteristic. 
The structured light method is fit to obtain the Lambert object and monochrome object's geometric information.
When a structured light method is selected for inspection, engineers should choose a suitable off-the-shelf product or build their scanning device.
For example, the device's field/depth of view should adapt to workpiece's size/height.
It's difficult for engineers to measure structured light method's accuracy for inspection.
In this section, we evaluate a self-built structured light device's performance via our metric; according to this metric, then we show how to check the coplanarity of the TypeC's pin needles.

\begin{figure}[H]
	\centering
	\includegraphics[width=\columnwidth]{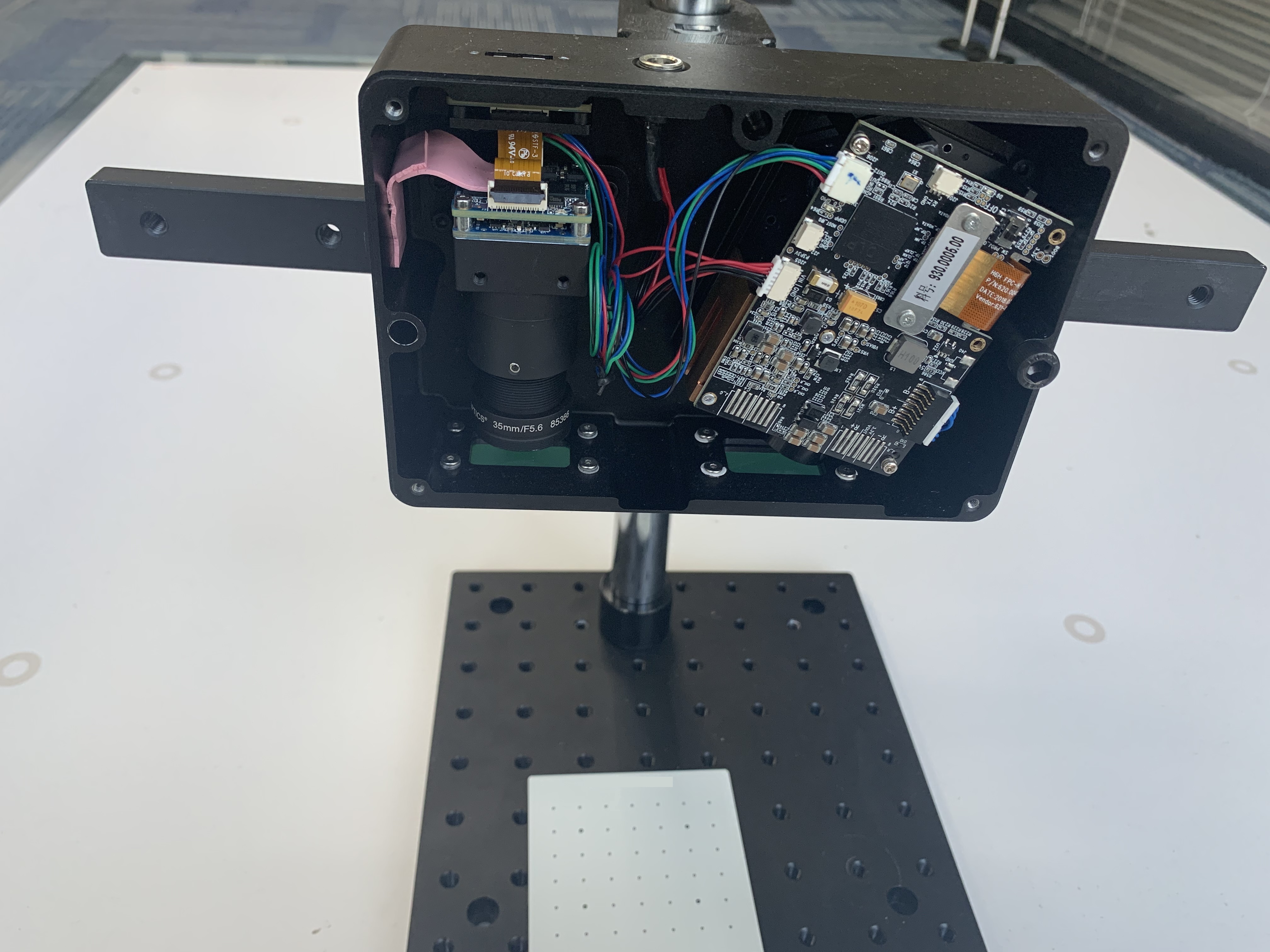}
	\caption{Our self-built structured light system which composed of one  camera and blue projector.}
	\label{setup}
\end{figure}
\begin{figure}[H]
	\centering
	\includegraphics[width=\columnwidth]{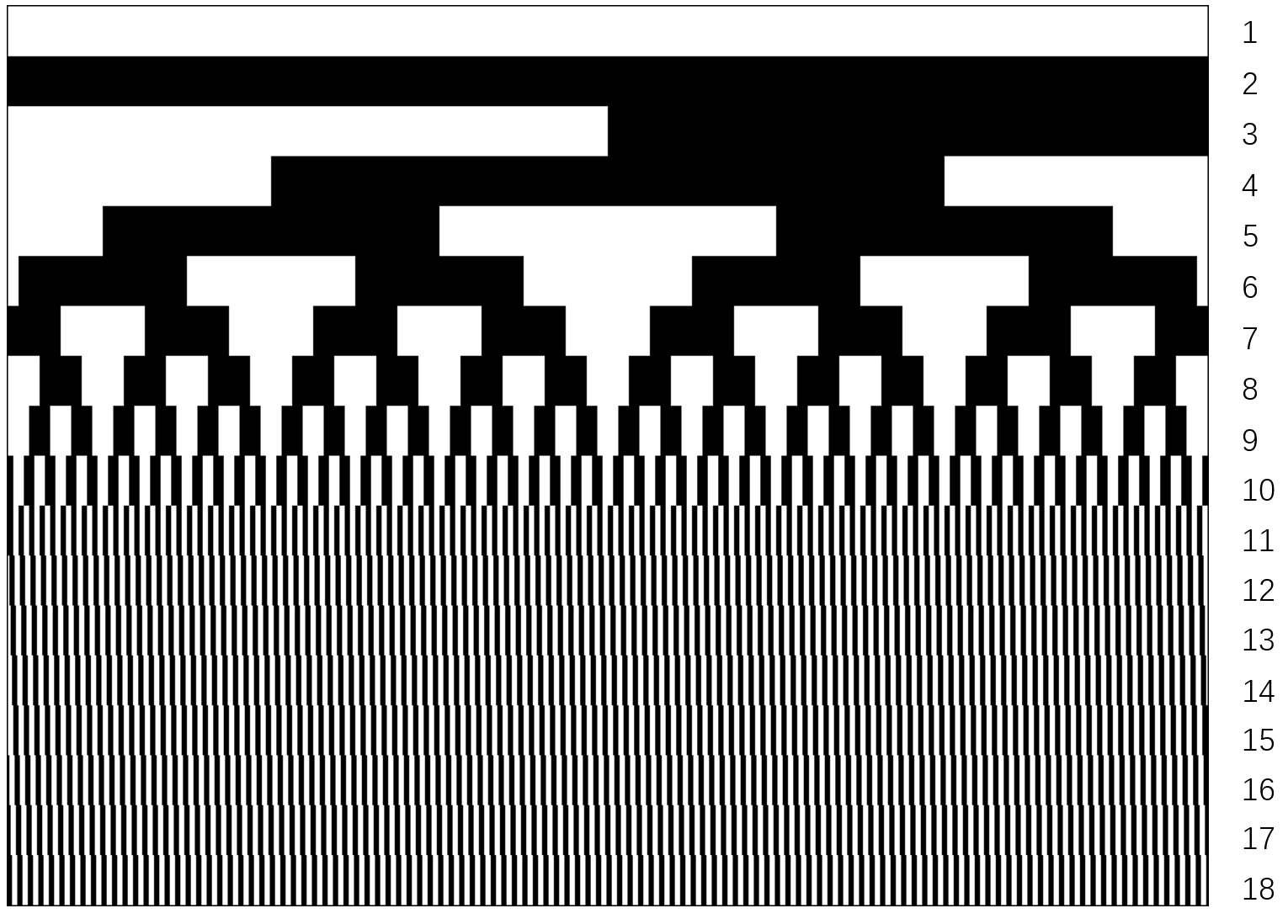}
	\caption{The coding strategy employs Gray code combined with strip shifting pattern.}
	\label{code}
\end{figure}

\begin{figure}[htb]
	\centering
	\includegraphics[width=\columnwidth]{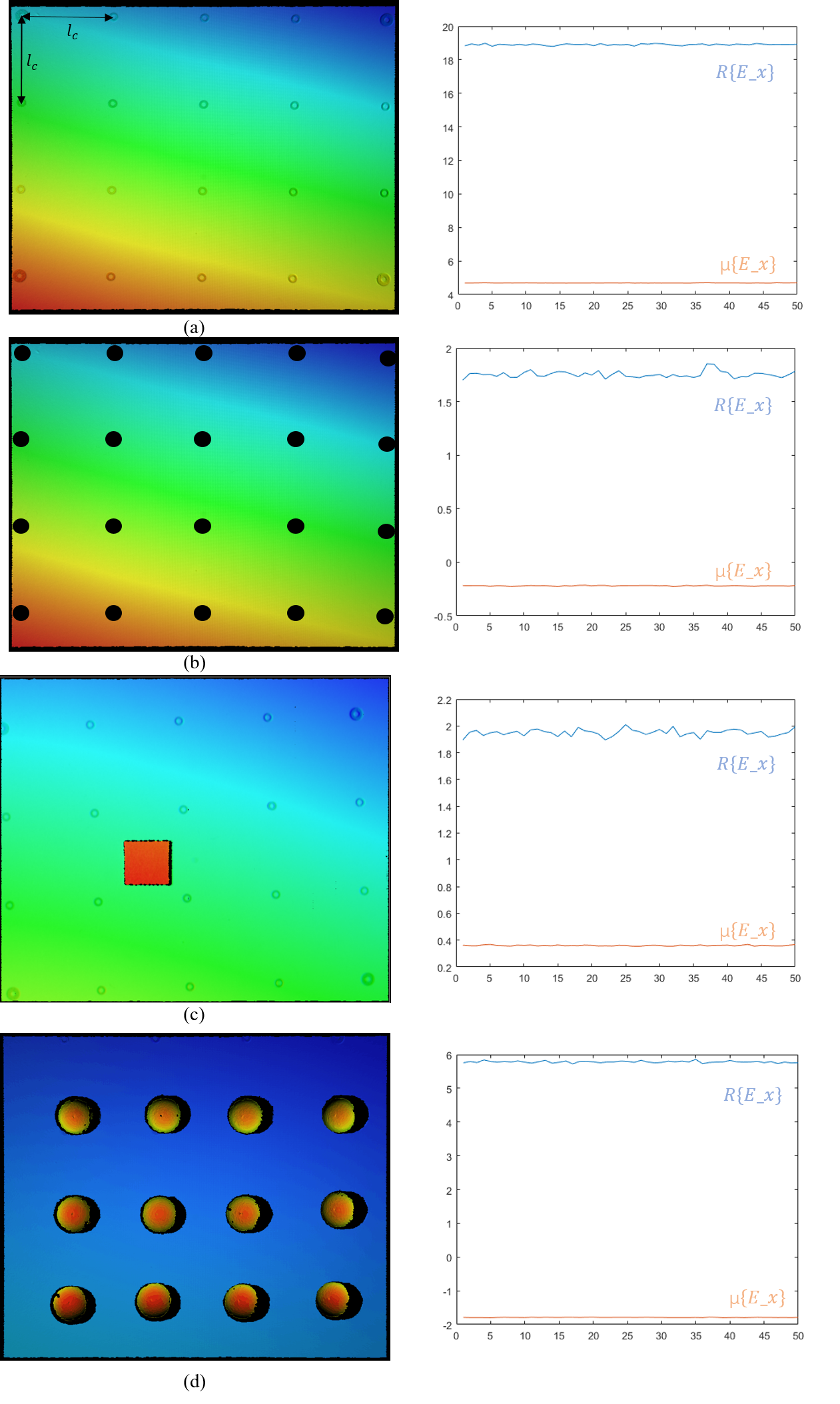}
	\caption{Evaluation metrics on our self-built structured light device.}
	\label{evs}
\end{figure}
To measure our metrics on structured light, we first built a structured light device with high accuracy.
The established structured light system is consists of one camera (FLIR BFS-U3-51S5M-BD, with the resolution of 2448$\times$2048, and maximum framerate up to 75Hz, pixel size is 3.45  \textmu m, USB3.1 interface) and one DLP projector (TI-DLP3010, a blue light source with a wavelength of 450nm, resolution 1280$\times$720, and maximum frame rates up to 4225Hz ), as shown in Fig. \ref{setup}.
As the above stated, the structured light method the industrial inspection adopted is often the time-multiplexing.
The coding strategy we adopted based on  \cite{zhansong2012} which employed the Gray code combined with line shifting pattern as shown in Fig. \ref{code}.
To further improve our device's accuracy, we adopted a practical calibration framework of \cite{RN87} to calibrate our system accurately. 

To evaluate our self-built device, a precise flat with several hollow circles markers, a gauge block and twelve tungsten steel balls is used as presented in Sec. 4.
As Fig. \ref{evs} shows,  Left column shows the captured point clouds of the evaluation tools, the right column shows the detailed data distribution of four metric's range and mean.
And Table \ref{t:t1} shows the four evaluation metrics' mean and standard variance on our self-built structured light device.

\begin{figure}[H]
	\centering
	\includegraphics[width=\columnwidth]{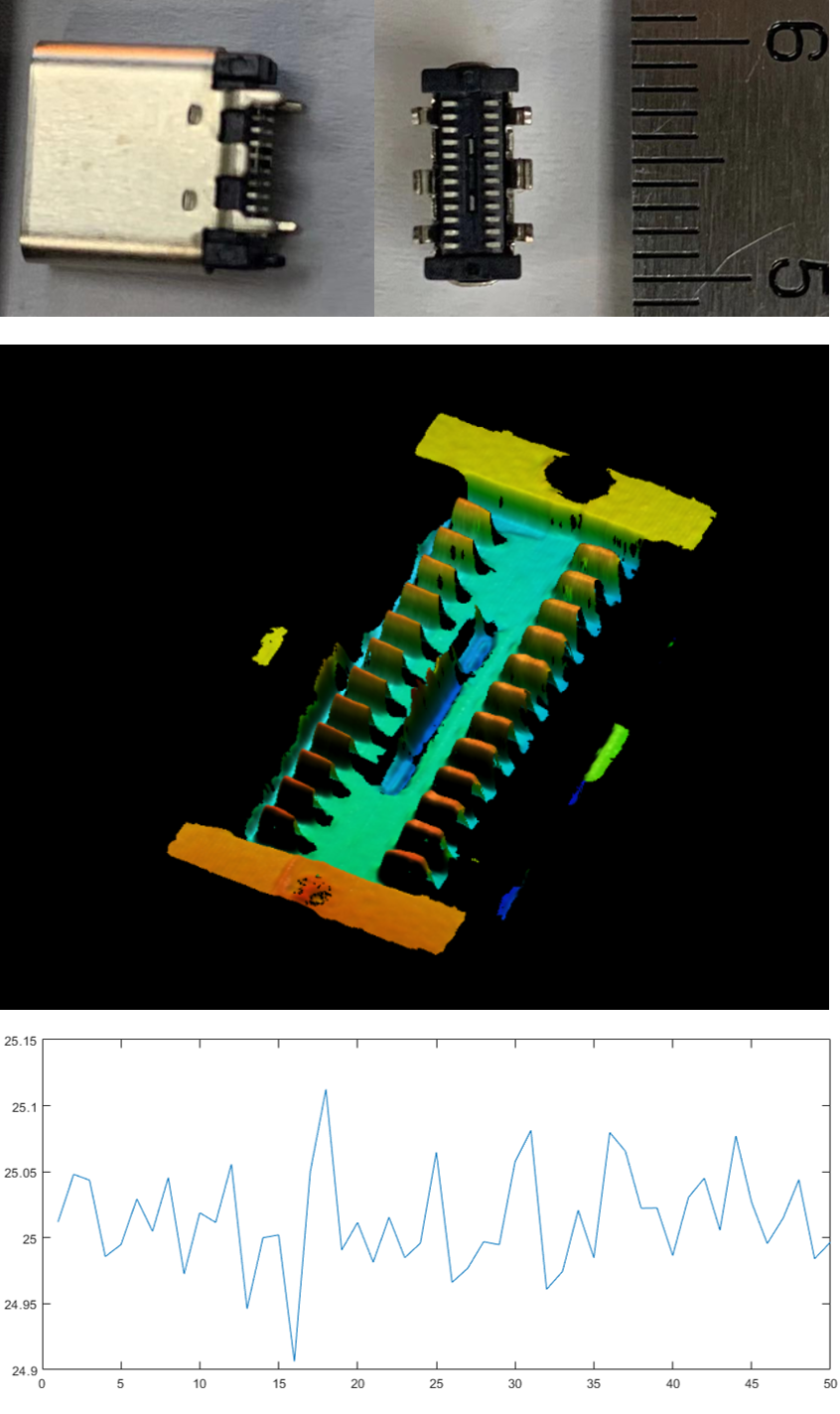}
	\caption{Top:TypeC's texture images, Middle:TypeC's point cloud images reconstructed by our self-built device,Bottom:the pin needles' coplanarity error range caculated via our device.}
	\label{pin24}
\end{figure}

And here comes an inspection sample of checking the coplanarity of the twenty-four pin needles of TypeC. 
A quality  TypeC product must satisfy that the maximum and minimum error from the fitting flat must be smaller than ten \textmu m.
By checking Table \ref{t:t1}, we can find the planarity and height metrics' range's mean far smaller than ten \textmu m and its standard variance is relatively small. 
So our self-built structured light can be integrated into the inspection machine for checking pin needles' coplanarity of TypeC.
To further prove that our metrics can evaluate the method/device accuracy for industrial inspection,
a non-qualified product with known range of 25 \textmu m was reconstructed by our self-built device; we can found that the range calculated is around the true value within 0.3 \textmu m shown in Fig.\ref{pin24} .
\begin{table}[H]
	\caption{Four evaluation metrics's mean and   standard variance on our self-built structured light device. }
	\centering
	\begin{tabular}{l|llll}
		\hline
		& $\mu(R(\{E\_x\}))$ & $\mu(R(\{E\_x\}))$ &$\mu(\mu(\{E\_x\}))$ & $\sigma(\mu(\{E\_x\}))$ \\ \hline
		E\_d  & 18.91 \textmu m  & 0.041    &  4.7 \textmu m   & 0.007    \\ \hline
		E\_p & 1.75   \textmu m & 0.029    & -0.22 \textmu m   & 0.003   \\ \hline
		E\_h & 1.95  \textmu m  & 0.025    & 0.36  \textmu m  & 0.004    \\ \hline
		E\_r & 5.78  \textmu m  & 0.031    & -1.79 \textmu m   & 0.006    \\ \hline
	\end{tabular}
	\label{t:t1}
\end{table}

\section{Conclusion}
In this paper, we put forward a  metric for evaluating the structured light method/device for industrial inspection.
First, we describe the mathematical model of laser and projector structured light. We then proposed an evaluation metric that can evaluate the structured light methods' performance on inspection tasks accurately.
The metric we proposed consists of four detailed criteria such as flatness, length, height and sphericity.
Each criterion is measured by the error's range/mean's data distribution.
In the experimental section, we evaluate a self-built device's performance via our metrics and present a real inspection case.  In the future, we are planning to measure the notable structured light method via our metric for helping the engineers to choose the suitable method/device quickly. How to incorporate our metric to improve calibration/reconstruction's accuracy is also another interesting work.
\bibliographystyle{cag-num-names}
\bibliography{refs}

\end{document}